\documentclass[preprint,12pt]{elsarticle}
\pdfoutput=1

\usepackage{amssymb}
\usepackage{graphicx}
\usepackage{caption}
\usepackage{epstopdf}
\usepackage{epsfig}
\usepackage{svg}
\usepackage{amsmath}
\usepackage{booktabs}
\usepackage{hyperref}
\hypersetup{colorlinks}

\journal{arXiv}
\begin{document}
	\bibliographystyle{elsarticle-num-names} 
	\begin{frontmatter}
		
		
		
		\title{ShapeEditer: a StyleGAN Encoder for Face Swapping}

		%
		
		\author{Shuai Yang, Kai Qiao}
		\address{Henan Key Laboratory of Imaging and Intelligent Processing, PLA Strategic Support Force Information Engineering University, Zhengzhou, China}
		%
		\begin{abstract}
			In this paper, we propose a novel encoder, called ShapeEditor, for high-resolution, realistic and high-fidelity face exchange. First of all, in order to ensure sufficient clarity and authenticity, our key idea is to use an advanced pretrained high-quality random face image generator, i.e. StyleGAN, as backbone. Secondly, we design ShapeEditor, a two-step encoder, to make the swapped face integrate the identity and attribute of the input faces. In the first step, we extract the identity vector of the source image and the attribute vector of the target image respectively; in the second step, we map the concatenation of identity vector and attribute vector into the $\mathcal{W+}$ potential space. In addition, for learning to map into the latent space of StyleGAN, we propose a set of self-supervised loss functions with which the training data do not need to be labeled manually. Extensive experiments on the test dataset show that the results of our method not only have a great advantage in clarity and authenticity than other state-of-the-art methods, but also reflect the sufficient integration of identity and attribute.
		\end{abstract}
		
		%
		
		\begin{keyword}
			Face Swapping, StyleGAN, Generative Adversarial Network
			
			
		\end{keyword}
		
	\end{frontmatter}
	
	
	
	\begin{figure}[h]
		\centering
		\includegraphics[scale=0.4]{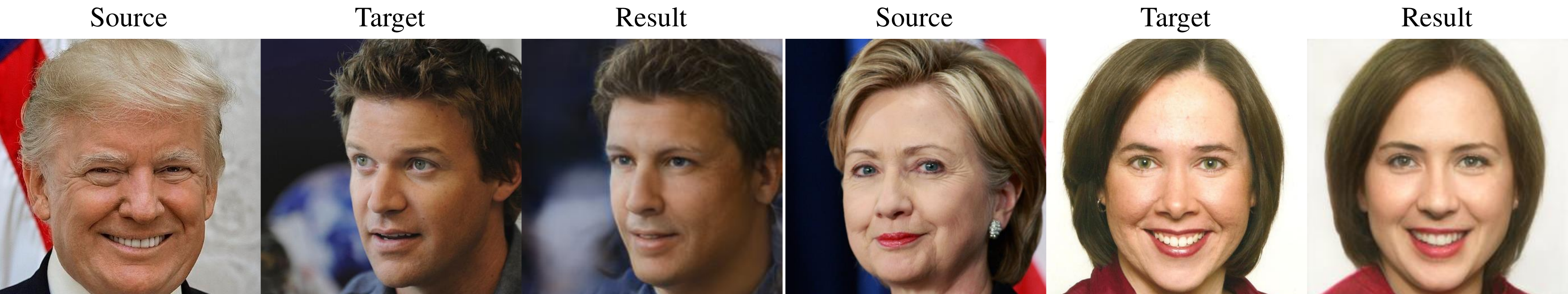}
		\caption{Transfer the identity of the source face to the target face. Results of ShapeEditor appear in the right. }
		\label{fig:net_structure}
	\end{figure}
	\section{Introduction}
	
	As one of the main contents of deepfake, face swapping declares to the world today that seeing is not always believing. Face swapping refers to transferring the identity of the source image to the face of another target image while keeping the illumination, head posture, expression, dress, background and other attribute information of the target image unchanged. Face swapping has received widespread attention since its birth, catering to the rich needs of social life, such as hairstyle simulation, film and television shooting, privacy protection\cite{ross2010visual} and so on.

	Face swapping is accompanied not only by its interesting and operational application prospects, but also by various challenges between reality and vision. The early face swapping methods\cite{bitouk2008face,korshunova2017fast} require a large number of images of source and target characters to provide sufficient facial information, otherwise the models will not have a suitable reference basis to convert reasonable results. Some 3D-based methods\cite{nirkin2018face,olszewski2017realistic,sun2018hybrid} make use of the advantage of fitting 3D face model to deal with the problems of large angle and small samples. At the same time, due to the limitation of the accuracy of 3D face model, it is impossible to generate works with more perfect details and higher fidelity. Recently, with the continuous tapping of the potential of Generative Adversarial Network (GAN), some face-swapping methods based on GAN\cite{natsume2018rsgan,bao2018towards,natsume2018fsnet,nirkin2019fsgan,li2019faceshifter} can achieve a good fusion of identity and attribute information with only a small number of samples, reflecting the effect of great creativity. Unfortunately, the surprising creativity of these methods does not offset the negative effects of their frequent artifacts and low resolution limitation.

	On another track, the most advanced face image generation methods have been able to generate facial images with high resolution and realistic texture. Most notably, StyleGAN\cite{karras2019style} can randomly generate a variety of clear faces with a resolution of up to $1024 \times 1024$. On this basis, some works\cite{richardson2020encoding,zhu2020domain,gu2020image,harkonen2020ganspace} deeply explore the  StyleGAN potential vector space; some\cite{shen2020interfacegan,shen2020closed,tewari2020stylerig} find a linear direction to control the change of a single facial attribute; some\cite{nitzan2020face} realize the control of facial expression and posture in the original StyleGAN image domain; some other\cite{richardson2020encoding,wang2021towards} perform well in dealing with the difficult task of facial super-resolution.

	We propose a many-to-many face swapping method based on pretrained StyleGAN\cite{karras2019style} model, which aims to ensure the clarity and fidelity of the results while fusing identity and attribute information. In view of the inherent ability of pretrained StyleGAN\cite{karras2019style} model to generate random high-quality face images, the difficulty of this task is how to accurately generate the corresponding latent vectors. In order to achieve this goal, we first design a novel encoder, ShapeEditor, to find the corresponding codes in the $\mathcal{W+}$ vector space. The work flow of the encoder is divided into two stages, the first is the  respective extraction of identity and attribute codes; the second is to map the combination of two-channel codes into the potential input vector domain of the pretrained model. Moreover, we design a set of loss functions with strong monitoring ability to urge ShapeEditor to update parameters so as to learn to map into the latent space of StyleGAN\cite{karras2019style} step by step. As verification, we have done a lot of qualitative and quantitative comparisons with the existing face swapping methods in the experiments, which shows the unique advantages of our method.

	In summary, the main contributions of this paper are: 
	\begin{itemize}
		\item \textbf{A new StyleGAN latent space encoder. } Our encoder can effectively mix two channels and map them into $\mathcal{W+}$ space of StyleGAN.
		\item \textbf{A novel face swapping method.} To the best of our knowledge, our method is the first to swap face based on StyleGAN, while having higher clarity and authenticity, comparable fidelity than other state-of-the-art works. 
	\end{itemize}
	
	\section{Related Works}
	Face swapping has always been a focus of attention, which is applied for privacy protection, entertainment and art, mainly using face alignment, facial key point detection, 3D face modeling, GAN and other technologies. Recently, StyleGAN\cite{karras2019style}, which has excellent performance in generating real and clear faces, has been proposed and further used in a variety of face editing operations.
	\subsection{Face Swapping}
	Bitouk et al.\cite{bitouk2008face} swap the face by replacing the pixels in the inner region, which is limited by the fact that the posture of the exchanged face must be the same. Korshunova et al.\cite{korshunova2017fast} use Convolutional Neural Networks (CNN) with multi-scale structures for face exchange and pose alignment with the assist of facial key points. Olszewski et al.\cite{olszewski2017realistic} fit the 3D face model of the source face, and use the Deep Neural Network (DNN) to infer the texture of the real mouth, with only a single RGB image. Sun et al.\cite{sun2018hybrid} train the CNN to regress the 3D model parameters of the input face, then replace the identity parameters, render the face image and combine the region around the head to generate a real face-swapping image. Limited to the accuracy of model reconstruction, the 3D-based face-swapping methods are unsatisfactory in terms of attribute and identity fidelity. Bao et al.\cite{bao2018towards} use two separate encoders to decouple the identity and attributes of human face, and propose an asymmetric loss training generation network, so that the generated results can better integrate the two pieces of information. Natsume et al.\cite{natsume2018rsgan} construct a transformation model consisting of two Variational Auto-Encoders (VAE) and a GAN, in which two VAEs extract the coding of hair and facial regions respectively, and GAN is responsible for reconstructing facial images. Nirkin et al.\cite{nirkin2019fsgan} divide face-swapping into three stages: reenactment, inpainting and blending, and use supervised methods to collect data and train four targeted generators. The model can effectively deal with facial hair occlusion, but it will fail in the face of facial occlusion such as glasses, veil and so on. Li et al.\cite{li2019faceshifter} use multi-level attribute encoders to extract attribute information, and adopt asymmetric training strategy similar to Bao et al.\cite{bao2018towards} to train identity and attribute integration networks, then train heuristic occlusion repair networks through self-supervision in the second stage to deal with multiple occlusion.
	\subsection{Latent Space of StyleGAN}
	Like other GANs with strong generating ability, StyleGAN has been studied by many researchers to understand and control its potential vector space since its birth. The inversion task of StyleGAN is to find the potential vector that best matches the given image. Abdal et al.\cite{abdal2019image2stylegan} directly adjust the $\mathcal{W+}$ vector by minimizing the error between the input and the output, thus embedding any face into the StyleGAN image domain, which often takes several minutes; some other works\cite{richardson2020encoding,zhu2020domain,gu2020image} choose to train the encoder to improve the inversion efficiency. Semantic control based on potential vector space is another widely concerned research direction. One way\cite{harkonen2020ganspace,shen2020interfacegan,shen2020closed} is to find a linear direction to change individual attribute (smile, age, facial angle etc.). In addition, Tewari et al.\cite{tewari2020stylerig} train the manipulation network to establish a close relationship between 3D semantic parameters and real facial expressions; Nitzan et al.\cite{nitzan2020face} realize the disentanglement of identity and expression, and mapped the spliced vectors to $\mathcal{W}$ space to achieve facial posture control in the limited original StyleGAN image domain; some other works\cite{richardson2020encoding,wang2021towards} realize the super-resolution of low-quality facial images with the help of StyleGAN. To the best of our knowledge, there is no research on face swapping based on StyleGAN, which is what we have done in this paper.
	
	\section{Method}
	\begin{figure}[ht]
		\centering
		\includegraphics[scale=0.5]{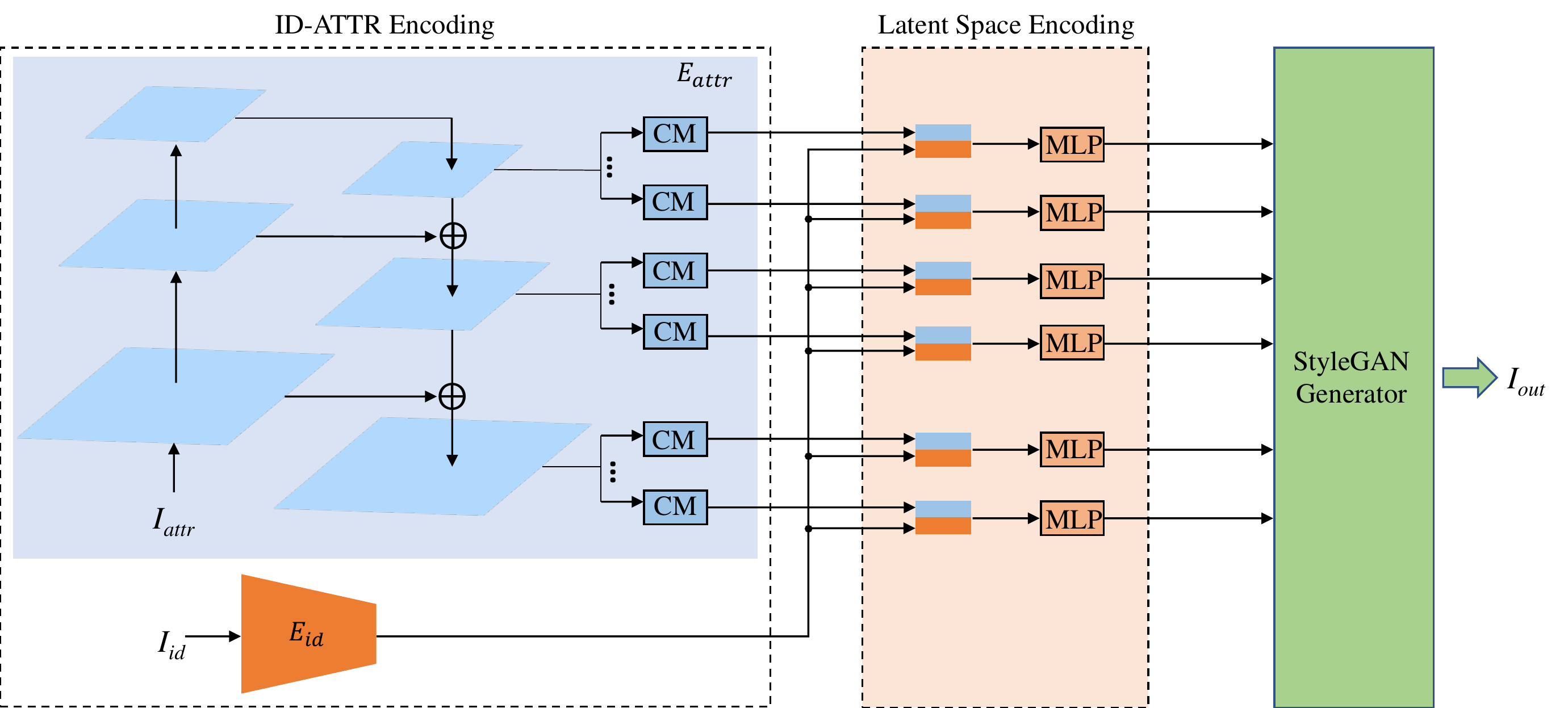}
		\caption{The flow of our method. There are two main steps: ID-ATTR Encoding and Latent Space Encoding. ID-ATTR Encoding extracts the attribute information of $I_{attr}$ and the identity information of $I_{id}$ using $E_{attr}$ and $E_{id}$ respectively, in which CM in $E_{attr}$ is a Convolutional Mapper; Latent Space Encoding further maps the splicing vector of identity and attribute to $\mathcal{W+}$ potential space to make the mixed information meet the input requirement of StyleGAN.}
		\label{fig:net_structure}
	\end{figure}
	Different from other works, the first criterion we pursue is that the images after face swapping have both resolution of up to $1024 \times 1024$ and high authenticity. In order to ensure the high quality of human face, our framework is based on facial pretrained StyleGAN\cite{karras2019style}, which has rich semantic expression ability. StyleGAN has three potential spaces: initial potential space $\mathcal{Z}$, intermediate potential space $\mathcal{W+}$ and extended potential space $\mathcal{W+}$. It is proved\cite{abdal2019image2stylegan} that the concatenation of 18 different 512-dimensional vectors is the easiest to embed the image and get a more reasonable result, so the operation space we choose is $\mathcal{W+}$.

	Our method requires two images as input: $I_{attr}$ and $I_{id}$. We expect the output of the model to reflect the identity of $I_{id}$, as well as the facial expression, head posture, hairstyle, lighting and other attribute information of $I_{attr}$. Therefore, the main challenge of this work is to obtain StyleGAN potential vectors which are consistent with the $\mathcal{W+}$ spatial distribution and better integrate attributes and identity. In order to solve this problem, we design a two-step coding process. As shown in Figure~\ref{fig:net_structure}, the entire mapping process is divided into two phases: ID-ATTR Encoding and Latent Space Encoding. In the first stage, $E_{id}$ extracts the identity vector of $I_{id}$, and $E_{attr}$ extracts the attribute vector of $I_{attr}$. Inspired by pSp\cite{richardson2020encoding}, $E_{attr}$ consists of a pyramid-shaped three-layer feature map extraction structure and a set of Convolutional Mappers (CM). In the second stage, we input the concatenation of $E_{id} (I_{id})$ and $E_{attr} (I_{attr})$ into the Multilayer Perceptron (MLP) of each layer, and map the vectors containing identity and attribute information directly to the $\mathcal{W+}$ potential vector space. In summary, the whole image conversion process can be represented as :
	\begin{equation}
		I_{out}=G(MLP([E_{id}(I_{id}),E_{attr}(I_{attr})])),
	\end{equation}
	where $G(\cdot)$ represents the pretrained StyleGAN\cite{karras2019style} model, $MLP(\cdot)$ represents the Multilayer Perceptron, $[\cdot,\cdot]$ represents the concatenation of two vectors.
	
	\subsection{Network Architecture}
	$E_{id}$ is pretrained ArcFace\cite{deng2019arcface} model. The Convolutional Mapper (CM) is a multi-layer convolution network with a fully connected module at the last layer. The three-layer feature map extraction structure is ResNet-50\cite{he2016deep}. MLP is a four-layer fully connected network. StyleGAN Generator is a pretrained model trained on FFHQ\cite{karras2019style}.
	\subsection{Loss Functions}
	The advanced face recognition model can accurately identify the face, so we believe that it can extract face feature information and take the feature vector extracted by pretrained ArcFace\cite{deng2019arcface} as the identity information. To ensure the identity of $I_{out}$ consistent with $I_{id}$, we introduce the identity loss :
	\begin{equation}
		\mathcal{L}_{id}=\Vert E_{id}(I_{id})-E_{id}(I_{out}) \Vert_2,
	\end{equation}
	where $E_{id}(\cdot)$ is the pretrained ArcFace\cite{deng2019arcface} model.

	Similarly, we need to adopt certain restriction to ensure that the attribute information of $I_{out}$ is consistent with that of $I_{attr}$.In view of the fact that the three-layer feature map extraction structure should gradually have the ability to extract attribute information with the training process, we define the attribute loss function:
	\begin{equation}
		\mathcal{L}_{attr}=\Vert P(I_{attr})-P(I_{out}) \Vert_2^2,
	\end{equation}
	where $P(\cdot)$ represents the extraction structure.

	We should note that the attribute information of $I_{attr}$ and the identity information of $I_{id}$ should not only exist in $I_{out}$, but also should be well integrated. Based on this idea, we define the reconstruction loss:
	\begin{equation}
		\mathcal{L}_{rec} =
		\begin{cases}
			\Vert I_{out}-I_{id} \Vert_2+\Vert F(I_{out})-F(I_{id}) \Vert_2
			& \text{if } I_{id}=I_{attr},\\
			0 & \text{Otherwise.}
		\end{cases} 
	\end{equation}
	where $F(\cdot)$ is the perceptual feature extractor in the LPIPS\cite{zhang2018unreasonable} loss, which can extract the perceptual information of the image at the high-dimensional level. $\mathcal{L}_2$ loss measures the difference between the two images at the pixel level. It is worth noting that $\mathcal{L}_{rec}$ has a value greater than 0 only when $I_{id}$ and $I_{attr}$ are the same, because only in this case should $I_{out}$ and $I_{id}$ (or $I_{attr}$) be so consistent that they are exactly the same, otherwise we cannot expect a similar comparison between the two images.

	Overall, our total training loss is the weighted sum of all the above losses:
	\begin{equation}
		\mathcal{L}_{total}=\lambda_{id}\mathcal{L}_{id}+\lambda_{attr}\mathcal{L}_{attr}+\lambda_{rec}\mathcal{L}_{rec}.
	\end{equation}
	
	\section{Experiments}
	\noindent\textbf{Implementation Details:} We use the FFHQ\cite{karras2019style} dataset as the training set, and the value of loss weights is set to $\lambda_{id}=0.5,\lambda_{attr}=0.1,\lambda_{rec}=1.$ The ratio of the training data with $I_{id}=I_{attr}$ and $I_{id} \neq I_{attr}$ is set to $2:1$. During the training, the network parameters of $E_{id}$ and StyleGAN Generator remain unchanged, and the weights of the rest are updated with iterations. In order to compare with other methods, we train the model with images of $256\times256$ resolution in this chapter, and the result of $1024\times1024$ resolution is given in Appendix A. This model needs to be trained on a single NVIDIA TITAN RTX for about two days with Ranger optimizer\cite{richardson2020encoding}, while batchsize is set to 8 and the learning rate is set to 0.0001.
	\begin{figure}
		\centering
		\includegraphics[scale=0.88]{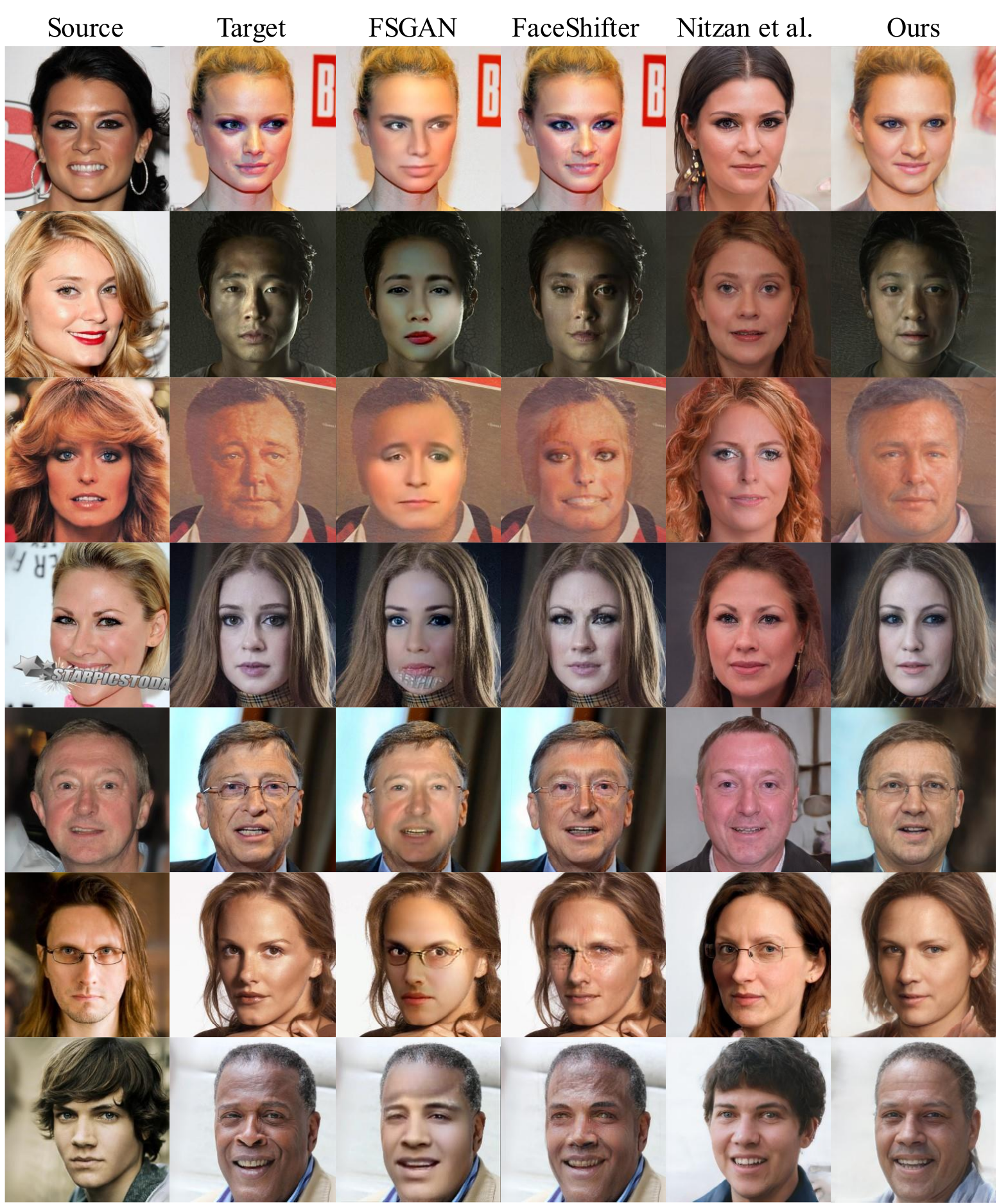}
		\caption{Qualitative comparison with FSGAN\cite{nirkin2019fsgan}, FaceShifter\cite{li2019faceshifter}, Nitzan et al.\cite{nitzan2020face} on the CelebAMask-HQ\cite{lee2020maskgan} test dataset. }
		\label{fig:comparision}
	\end{figure}

	We compare our method with FSGAN\cite{nirkin2019fsgan}, FaceShifter\cite{li2019faceshifter}, Nitzan et al.\cite{nitzan2020face} on the CelebAMask-HQ\cite{lee2020maskgan} test dataset. As shown in Figure~\ref{fig:comparision}, as expected, because our method is based on a pretrained StyleGAN\cite{karras2019style} with high-quality face generation capabilities, all the generation results (Figure~\ref{fig:comparision}, column 6) are stable and clear enough that there are no errors such as artifacts and abnormal illumination.

	Almost every output image (Figure~\ref{fig:comparision}, column 3) of FSGAN\cite{nirkin2019fsgan} shows unnatural lighting transition and lack of facial details, and the abnormal region of the face is caused by directly extracting and filling the internal region of the face (Figure~\ref{fig:comparision}, row 3, column 4), which is completely avoided in our method.

	Because there is no pretrained model as backbone, it is difficult for FaceShifter\cite{li2019faceshifter} to avoid facial blur, and some results even show facial illumination confusion (Figure~\ref{fig:comparision}, row 3, column 4) and eye ghosting (Figure~\ref{fig:comparision}, row 7, column 4), showing that its authenticity is far behind our method.

	Similar to our method, Nitzan et al.\cite{nitzan2020face} uses StyleGAN\cite{karras2019style} as backbone. But it can not accurately integrate identity and attribute information because of its simple encoder structure and the constraint of  $\mathcal{W}$ potential space. Therefore, although it can generate high-quality images (Figure~\ref{fig:comparision}, column 6), it is not as good as our method in semantic information fusion, which is reflected in that the attributes of the target image, such as hairstyle and background, are not contained.

	In addition to the excellent performance in terms of authenticity and fidelity, our method can also deal with extreme lighting condition (Figure~\ref{fig:comparision}, row 2, column 6) and even keep the sense of age (Figure~\ref{fig:comparision}, row 3, column 6). Thanks to that we use the facial recognition module to extract the identity vector instead of directly utilizing the pixels in the facial area, we can extract the identity information very well even if the source image has facial occlusion (Figure~\ref{fig:comparision}, row 4, column 6); our model understands the knowledge that whether its output should have glasses (Figure~\ref{fig:comparision}, column 6, rows 5-6), which is embedded in the potential space of the pretrained StyleGAN\cite{karras2019style} model.

	\section{Conclusion}
	In this paper, we propose a new face-swapping framework, including ShapeEditor and pretrained StyleGAN model. The pretrained model gives our framework the potential to generate clear and real faces, while the ShapeEditor encoder effectively extracts and integrates the attribute and identity information of the input images, and accurately maps them into the $\mathcal{W+}$ space, thus controlling the StyleGAN to output appropriate results. Extensive experiments show that our method performs better than other existing frameworks.
	\newpage
	\section*{References}
	\bibliography{style_reference}
\end{document}